\documentclass[sigconf]{acmart}
\AtBeginDocument{%
  \providecommand\BibTeX{{%
    \normalfont B\kern-0.5em{\scshape i\kern-0.25em b}\kern-0.8em\TeX}}}

\copyrightyear{2023}
\acmYear{2023}
\setcopyright{rightsretained}
\acmConference[MM '23]{Proceedings of the 31st ACM International Conference on Multimedia}{October 29-November 3, 2023}{Ottawa, ON, Canada}
\acmBooktitle{Proceedings of the 31st ACM International Conference on Multimedia (MM '23), October 29-November 3, 2023, Ottawa, ON, Canada}
\acmDOI{10.1145/3581783.3611775}
\acmISBN{979-8-4007-0108-5/23/10}

\usepackage[outdir=./]{epstopdf}

\makeatletter
\gdef\@copyrightpermission{
 \begin{minipage}{0.3\columnwidth}
  \href{https://creativecommons.org/licenses/by/4.0/}{\includegraphics[width=0.90\textwidth]{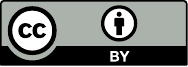}}
 \end{minipage}\hfill
 \begin{minipage}{0.7\columnwidth}
  \href{https://creativecommons.org/licenses/by/4.0/}{This work is licensed under a Creative Commons Attribution International 4.0 License.}
 \end{minipage}
 \vspace{5pt}
}
\makeatother

\usepackage{subfigure}
\usepackage{algorithmicx,algorithm}
\usepackage[noend]{algpseudocode}
\usepackage{color}
\usepackage{url}
\usepackage{balance}
\usepackage{multirow}
\usepackage{indentfirst}
\usepackage{booktabs}
\usepackage{amsmath}

\acmSubmissionID{361}

\begin{document}

\title{Speech-Driven 3D Face Animation with \\
Composite and Regional Facial Movements
}





\author{Haozhe Wu}
\email{wuhz19@mails.tsinghua.edu.cn}
\affiliation{%
  \institution{Department of Computer Science and Technology, Tsinghua University}
  \city{Beijing 100084}
  \country{China}
}

\author{Songtao Zhou}
\email{zhoust19@mails.tsinghua.edu.cn}
\affiliation{%
  \institution{Department of Computer Science and Technology, Tsinghua University}
  \city{Beijing 100084}
  \country{China}
}

\author{Jia Jia}
\email{jjia@tsinghua.edu.cn}
\authornote{Corresponding author.}
\affiliation{%
  \institution{Department of Computer Science and Technology, Tsinghua University}
  \institution{Beijing National Research Center for Information Science and Technology}
  \city{Beijing 100084}
  \country{China}
}

\author{Junliang Xing}
\email{jlxing@tsinghua.edu.cn}
\affiliation{%
  \institution{Department of Computer Science and Technology, Tsinghua University}
  \city{Beijing 100084}
  \country{China}
}

\author{Qi Wen}
\email{2838158073@qq.com}
\affiliation{%
  \institution{ByteDance}
  \city{Hangzhou}
  \country{China}
}

\author{Xiang Wen}
\email{wenxiang@zju.edu.cn}
\affiliation{%
  \institution{ByteDance}
  \city{Hangzhou}
  \country{China}
}

\renewcommand{\shortauthors}{Haozhe Wu et al.}

\begin{abstract}
Speech-driven 3D face animation poses significant challenges due to the intricacy and variability inherent in human facial movements. %
This paper emphasizes the importance of considering both the composite and regional natures of facial movements in speech-driven 3D face animation. %
The composite nature pertains to how speech-independent factors globally modulate speech-driven facial movements along the temporal dimension. Meanwhile, the regional nature alludes to the notion that facial movements are not globally correlated but are actuated by local musculature along the spatial dimension. It is thus indispensable to incorporate both natures for engendering vivid animation. To address the composite nature, we introduce an adaptive modulation module that employs arbitrary facial movements to dynamically adjust speech-driven facial movements across frames on a global scale. To accommodate the regional nature, our approach ensures that each constituent of the facial features for every frame focuses on the local spatial movements of 3D faces. Moreover, we present a non-autoregressive backbone for translating audio to 3D facial movements, which maintains high-frequency nuances of facial movements and facilitates efficient inference. Comprehensive experiments and user studies demonstrate that our method surpasses contemporary state-of-the-art approaches both qualitatively and quantitatively.

\end{abstract}




\maketitle

\section{Introduction}



Speech-driven 3D face animation is a crucial technology in the field of digital avatar synthesis, which has wide-ranging applications in VR/AR, games, and film-making. However, the intricacy and variability of human facial movements pose significant challenges for this task. %
Human facial movements have two essential natures: composite and regional. %
The composite nature refers to how speech-independent factors, such as talking styles and expressions, globally modulate speech-driven facial movements along the temporal dimension. For example, different talking styles can affect the amplitude of mouth opening while speaking the same word. %
The regional nature arises because the movement of different facial parts is not globally connected along the spatial dimension but rather determined by the action of local muscles. %
For instance, the movements of the eyebrows are usually uncorrelated with those of the jaw. Understanding and modeling composite and regional natures are crucial for realistic and vivid facial animation. %

\begin{figure*}[h]
  \centering
  \includegraphics[width=\linewidth]{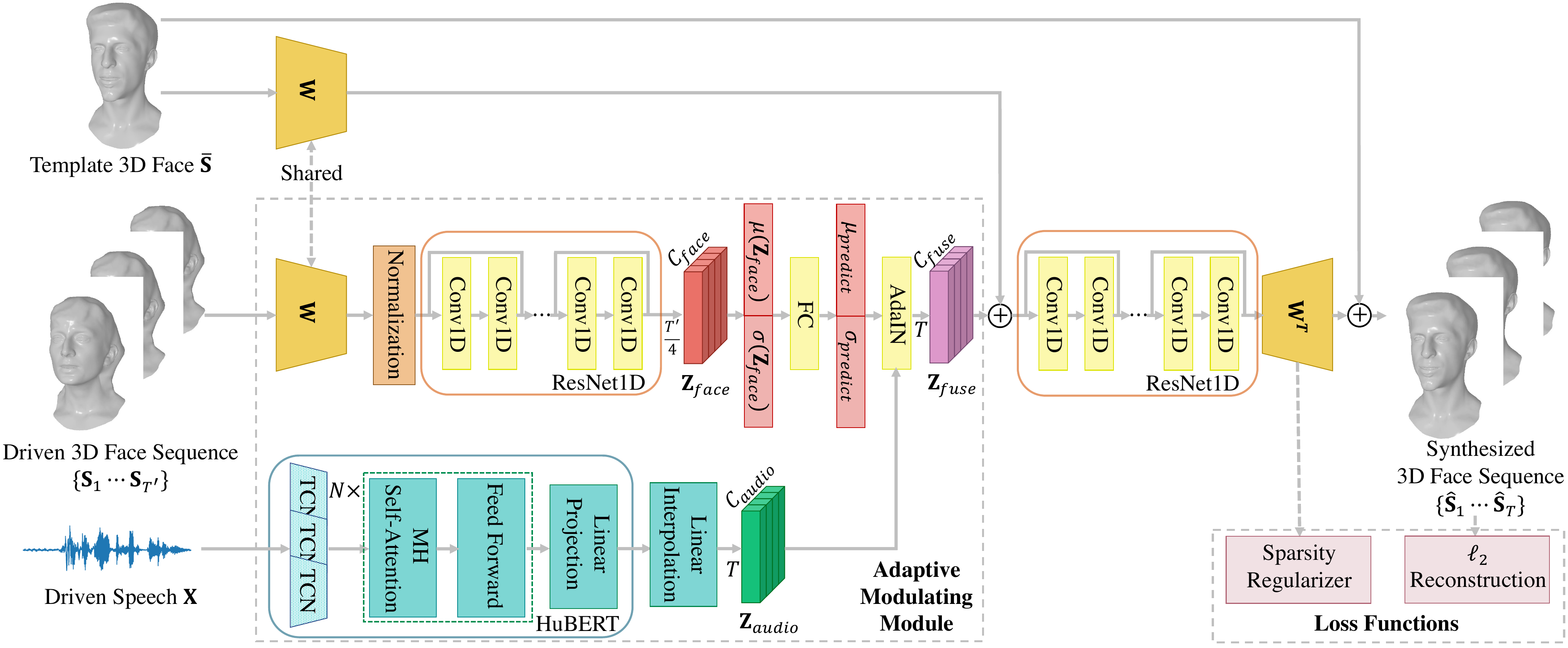}
  \caption{The overall framework of our method. The adaptive modulating module incorporates the composite nature of facial movements into the framework, while the sparsity regularizer interprets the regional nature of facial movements. %
  The overall backbone is non-autoregressive, which enables efficient training and inference.}
  \label{fig:framework}
\end{figure*}

Several previous studies~\cite{fan2022faceformer, xing2023codetalker, thambiraja2022imitator, gururani2022spacex} have attempted to incorporate the composite nature into speech-driven 3D face animation models by fusing speech-independent labels such as emotion, identity, and style. However, these fused labels are often coarse-grained, which limits their capacity to capture intricate interactions between speech-independent factors and speech-driven movements. To achieve fine-grained control over speech-independent factors, some approaches~\cite{richard2021meshtalk, peng2023emotalk, ma2023styletalk} have proposed disentangling speech-driven and speech-independent movements in a single 3D face sequence. However, these methods tend to oversimplify or only focus on local aspects of the composite nature, reducing the expressiveness of facial animations. %
Moreover, previous learning-based methods tend to overlook the regional nature of facial movements, while rule-based methods~\cite{edwards2016jali,xu2013practical,taylor2012dynamic,wu2006real} consider such nature but require extensive manual labor when animating unseen faces. %
To address the issues above, there is a critical need to develop a comprehensive method that captures a global understanding of the composite nature and considers the regional nature. %

To tackle these challenges, we propose a novel speech-driven 3D face animation method that considers both facial movements' composite and regional natures. We introduce an adaptive modulating module to account for the composite nature. %
This module inputs the latent audio features and arbitrary 3D face sequence, extracting global-aware speech-independent representations and modulating the latent audio features according to the extracted representations. %
To accommodate the regional nature, we propose a sparsity regularizer, which, for each frame, enforces each facial feature element to focus on the local region of mesh vertices. %
Furthermore, we present a non-autoregressive backbone for translating audio to 3D facial movements. %
We apply the pretrained HuBERT model~\cite{hsu2021hubert} to extract high-level audio features and adopt ResNet~\cite{he2016deep} with 1D convolution to serve as the motion decoder. %
The overall framework is shown in Figure~\ref{fig:framework}. %
Our backbone both enables efficient inference and preserves high-frequency motion details. %

To demonstrate the effectiveness of our framework, we conduct extensive experiments on the VOCA~\cite{cudeiro2019capture}, MeshTalk~\cite{richard2021meshtalk}, and BIWI~\cite{fanelli20103} datasets. %
We evaluate our framework against several state-of-the-art approaches using various quantitative metrics, including the lip vertex error, face error, and dynamic time wrapping error. Moreover, we conduct a user study to evaluate the naturalness
of facial movements and the synchronization between speech and animation. The experimental results show that our proposed framework outperforms existing methods in terms of both quantitative metrics and subjective evaluations, verifying that it is beneficial to consider both composite and regional natures in the task of speech-driven 3D face animation. %
The code is publicly available at \href{https://github.com/wuhaozhe/audio2face\_mm2023}{https://github.com/wuhaozhe/audio2face\_mm2023}. %



\section{Related Work}







Speech-driven face animation has received significant attention in previous literature. %
Considerable research has focused on animating 2D faces~\cite{chen2018lip,das2020speech,fan2015photo,ji2021audio,prajwal2020lip,vougioukas2020realistic,zhou2019talking,sinha2022emotion,ji2022eamm,gururani2022spacex,huang2022audio,alghamdi2022talking}, while we concentrate on animating 3D faces in this work. %
In 3D facial animation, rule-based methods have been previously explored \cite{edwards2016jali,xu2013practical,taylor2012dynamic,wu2006real}. These methods rely on breaking down facial movements into smaller units, such as visemes \cite{mattheyses2015audiovisual} and facial action units (FAUs) \cite{ekman1978facial}, and establishing mappings between speech and these units. These rule-based methods take into account the regional nature of facial movements, with both visemes and FAUs designed based on anatomical characteristics of the human face. As a result, they have achieved satisfactory results in synthesizing lip motions. However, these methods require extensive manual labor when animating unseen faces and their performance is limited in synthesizing speech-independent movements, such as facial expressions that are not directly related to speech. %

With the advent of 4D face datasets~\cite{richard2021meshtalk,cudeiro2019capture,fanelli20103,wu2023mmface4d}, various learning-based methods have emerged~\cite{hwang2022audio}.
A few methods focus on driving one particular character~\cite{karras2017audio,zhou2018visemenet}, %
while most methods work on driving different identities~~\cite{fan2022faceformer,xing2023codetalker,thambiraja2022imitator,gururani2022spacex,peng2023emotalk,richard2021meshtalk,ma2023styletalk,ji2021audio,tang2022memories,ye2023geneface,wu2021imitating,yi2020audio}. %
In these methods, researchers have attempted to consider the composite nature. %
Some methods regard speech-independent factors as one-hot labels. %
For example, methods such as FaceFormer~\cite{fan2022faceformer} and CodeTalker~\cite{xing2023codetalker} treated the one-hot identity label as speech-independent factors and fused identity label with speech audio in the Transformer~\cite{vaswani2017attention} decoder. %
These two methods synthesize discrete talking styles for each identity and enable interpolation between different styles. However, they are limited to synthesizing unseen talking styles, which poses a challenge for their practical applications. %
In the footsteps of the FaceFormer and CodeTalker methods, the Imitator~\cite{thambiraja2022imitator} approach introduced a style adaptive motion decoder, which allows for fine-tuning on previously unseen styles. Nonetheless, this fine-tuning process incurs a significant computational cost and hampers fast generalization. %
In addition to identity labels, the SpaceX method~\cite{gururani2022spacex} also incorporates one-hot emotion labels to generate expressive facial animations. %
Although these methods that utilize one-hot speech-independent labels can produce diverse facial movements, the representation granularity of speech-independent factors is often too coarse, which hinders their broader application.

\begin{figure}[t]
  \centering
  \includegraphics[width=\linewidth]{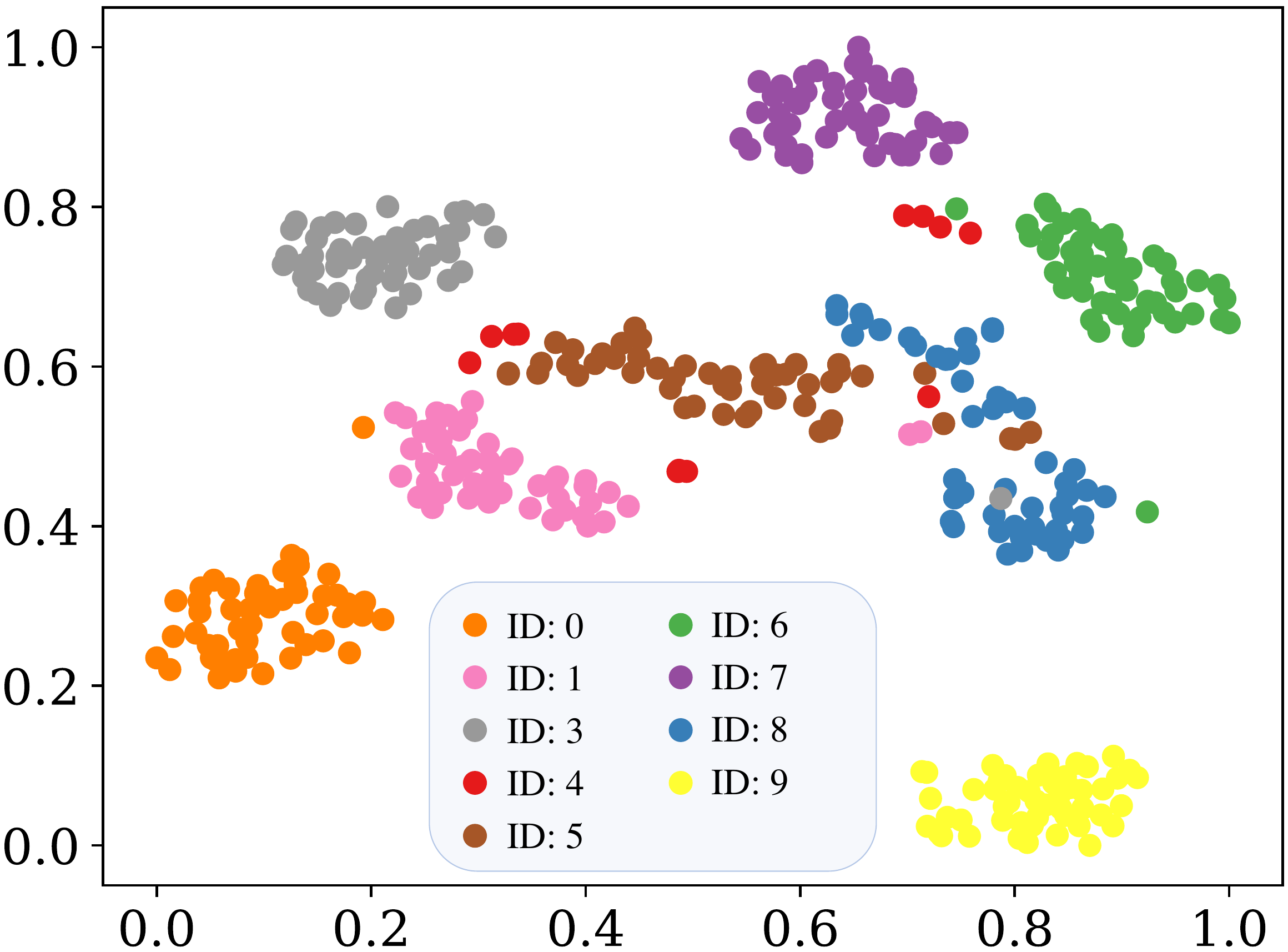}
  \caption{The t-SNE visualization of how speech-independent factors influence facial movement distributions. %
  Each point represents the statistics of one 3D sequence. Points belonging to different identities are colorized with different colors.}
  \label{fig:composite}
\end{figure}

To achieve fine-grained control over speech-independent factors, several methods~\cite{richard2021meshtalk,peng2023emotalk,ma2023styletalk,tang2022memories,ye2023geneface,wu2021imitating,ji2021audio} have proposed to disentangle speech-driven and speech-independent movements in a single 3D face sequence. %
Some of these methods achieve disentanglement in a supervised manner~\cite{richard2021meshtalk,peng2023emotalk,ji2021audio}, such as the MeshTalk~\cite{richard2021meshtalk} method, which uses a cross-modality loss to disentangle the speech-driven and speech-independent facial movements. %
However, the assumption made by MeshTalk that speech-driven and speech-independent movements correspond respectively to the lower and upper parts of human faces may not hold for all facial expressions or movements. %
In contrast, the Emotional Video Portraits (EVP)~\cite{ji2021audio} and EmoTalk~\cite{peng2023emotalk} methods improve on MeshTalk by disentangling speech emotion and speech content through cross-reconstruction without making such assumptions. Nonetheless, these methods still require one sentence to be spoken with different emotions, which poses challenges for dataset collection. %
Some methods achieve disentanglement in an unsupervised manner~\cite{tang2022memories,ye2023geneface,wu2021imitating,ma2023styletalk}, the MemFace method~\cite{tang2022memories} uses a latent memory dictionary to disentangle speech-independent factors. In contrast, the GeneFace~\cite{ye2023geneface} method incorporates uncertainty into the synthesis model to represent speech-independent movements statistically. In addition, some methods~\cite{ma2023styletalk,wu2021imitating} directly extract speech-independent representations from the reference video and fuse such representations with audio embeddings. %
To conclude, while the aforementioned methods have shown promising results in modeling speech-independent movements, they tend to oversimplify or only focus on the local aspects of the composite nature. %
The global understanding of the composite nature is lacking in these methods. %
Moreover, these learning-based methods neglect the regional nature of facial movements, further hindering the ability to synthesize realistic and vivid animations. These limitations highlight the need for a comprehensive approach that considers both facial movements' composite and regional natures for improved 3D facial animation synthesis. %





\begin{figure}[t]
  \centering
  \includegraphics[width=\linewidth]{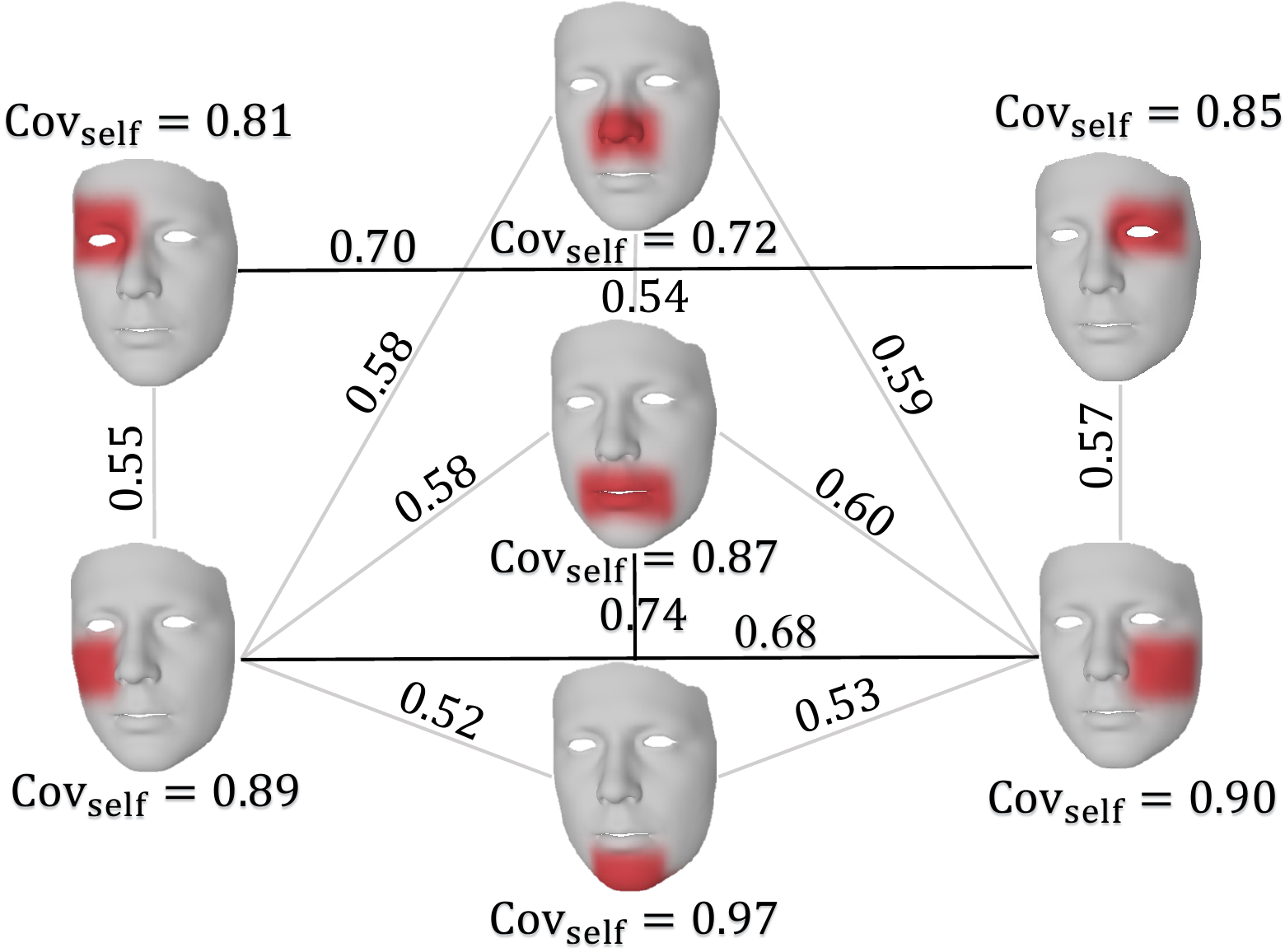}
  \caption{The correlation graph of local facial regions. The local facial regions are colorized red. %
  The $\mathrm{Cov_{self}}$ denotes the self-correlation inside the local region, and the weight of the edge denotes the correlation between the two regions.}
  \label{fig:regional}
\end{figure}

\section{Observations of 3D Face Animation}
In this section, we systematically investigate the impact of the composite and regional natures on the 3D face animations. To this end, we conduct data observations for each nature separately. %

We explore the impact of the composite nature by visualizing how speech-independent factors influence the distribution of facial movements. 
Specifically, we calculate the standard deviation for each vertex of a 3D sequence along the temporal dimension, then reduce the standard deviations of all 3D mesh vertices to 2D using t-SNE~\cite{van2008visualizing}, which we visualize as 2D points. %
Figure~\ref{fig:composite} demonstrates the visualization results. Different colors are used to plot points belonging to different identities. Our findings suggest that the speech-independent factor such as speaker identity significantly impacts the distribution of facial movements.

To investigate the impact of the regional nature on facial movements, we analyze motion correlation across different regions of the face. Motion correlation is used to quantify the degree of dependency between the movements of different facial regions. We first partition facial movements into small regions based on the facial mesh vertices to obtain the motion correlation. We then calculate the pairwise correlation coefficient along the spatial dimension between each pair of regions, resulting in a correlation matrix. Finally, we visualize this matrix as a connected graph, where edges with less than 0.5 correlation coefficient are removed. %
Figure~\ref{fig:regional} shows the correlation graph. %
The correlation graph reveals that the motion correlation of facial regions is not uniformly distributed, with some regions being more correlated than others. 
For example, the mouth and chin regions, which are responsible for similar expressions, display a higher correlation, while the upper and lower parts of the face have a lower correlation.

These findings have important implications for understanding the underlying mechanisms of 3D facial movements. %
In the next section, we propose a framework that comprehensively considers both composite and regional natures. %


\section{Methodology}

As mentioned earlier, the composite and regional natures have a significant impact on facial movements, and it is crucial to take them into account when generating realistic and accurate 3D face animations. %
In this section, we detailedly elaborate on how we integrate these two natures into the speech-driven 3D face animation framework. %
Moreover, we introduce our non-autoregressive backbone, which preserves high-frequency details of facial animations and enables efficient inference. %

\subsection{Problem Formulation}

Before introducing the overall framework, we first formulate the problem of audio-driven 3D face animation in the presence of speech-independent factors. %
This task takes three inputs: a template 3D face $\mathbf{\bar{S}}$ of the target person, the driven speech $\mathbf{X}$ with duration $T$, and the driven 3D face sequence $\{\mathbf{S}_{1} \cdots \mathbf{S}_{T^{'}}\}$. %
The objective is to synthesize a sequence of 3D face animations $\{\mathbf{\hat{S}}_{1} \cdots \mathbf{\hat{S}}_{T}\}$. The synthesized animations have the same identity as $\mathbf{\bar{S}}$, are synchronized to the driven speech $\mathbf{X}$, and incorporate the speech-independent facial movements of $\{\mathbf{S}_{1} \cdots \mathbf{S}_{T^{'}}\}$. %

It is worth noting that we do not synthesize the blendshape weights of 3D faces but rather directly synthesize the 3D face vertices. %
There are two main reasons for this choice. %
Firstly, blendshapes often result in a loss of high-frequency facial information, whereas the 3D face sequence preserves all facial details. By synthesizing 3D face sequences directly, our model can capture intricate facial movements and fine-grained nuances. %
Secondly, the blendshape weight is often limited by its uninterpretable definitions, whereas directly animating 3D faces has broader applications. %
Based on such a setting, all of the input 3D faces and the synthesized 3D faces have a shape of $N \times 3$, where $N$ is the number of mesh vertices. %
For different datasets, the vertex number $N$ is different. %

\subsection{Adaptive Modulating Module}

To incorporate the composite nature in 3D facial animation synthesis, it is essential to effectively combine both speech-independent and speech-driven facial movements. %
To achieve this, we propose the adaptive modulating module, which plays a critical role in effectively blending these two types of movements. This module first extracts global-aware speech-independent representations that capture the facial movements not influenced by the speech signal. These representations are then used to modulate the latent audio features, allowing the model to dynamically adjust the contribution of speech-driven and speech-independent factors to each specific facial region. By utilizing the adaptive modulating module, our framework synthesizes more diverse and natural face animations.

Figure~\ref{fig:framework} shows the adaptive modulating module. %
The module extracts the speech-independent representations from $\{\mathbf{S}_{1} \cdots \mathbf{S}_{T^{'}}\}$. %
More specifically, for each 3D face $\mathbf{S}_{i}$, we first normalize $\mathbf{S}_{i}$ by subtracting the mean face of $\{\mathbf{S}_{1} \cdots \mathbf{S}_{T^{'}}\}$. %
Formally:
\begin{equation}
    \mathrm{Norm}(\mathbf{S}_{i}) = \mathbf{S}_{i} - \frac{\sum_{i=1}^{t}\mathbf{S}_{i}}{t}.
\end{equation}
The goal of normalization is to extract solely the facial movement information while removing the identity information. Subsequently, we reduce $\mathrm{Norm}(\mathbf{S}_{i})$ to a low-dimensional feature vector with embedding matrix $\mathbf{W}$. We then concatenate the embedded vector of the 3D faces into a sequence and input this sequence into a ResNet1D~\cite{he2016deep} encoder, thereby obtaining the latent face representations $\mathbf{Z}_{face}$. %
$\mathbf{Z}_{face}$ has a shape of $\frac{T^{'}}{4} \times C_{face}$, where $C_{face}$ is the channel number, and $\frac{T^{'}}{4}$ accounts for the downsampled embedded face vectors along the temporal dimension. %
Afterward, we extract the speech-independent facial movements from $\mathbf{Z}_{face}$. %
Different from the previous methods~\cite{richard2021meshtalk,peng2023emotalk,ji2021audio} that leverages cross reconstruction loss to extract speech-independent factors, 
we extract the speech-independent information by simply calculating the mean $\mu(\cdot)$ and standard deviation $\sigma(\cdot)$ of $\mathbf{Z}_{face}$ along the temporal dimension. %
Remarkably, this simple approach provides an effective representation of speech-independent information, as it captures the overall statistical distribution of face animations while excluding the temporal information of $\mathbf{Z}_{face}$. %

Having obtained $\mu(\mathbf{Z}_{face})$ and $\sigma(\mathbf{Z}_{face})$, %
we now blend them with input speech signals. %
We first feed the input speech to the pretrained audio model~\cite{hsu2021hubert}, yielding the latent audio feature $\mathbf{Z}_{audio}$ with a shape of $T \times C_{audio}$, where $C_{audio}$ is the channel number of latent audio features. %
Afterwards, $\mu(\mathbf{Z}_{face})$ and $\sigma(\mathbf{Z}_{face})$ are used to modulate the mean and standard deviation of $\mathbf{Z}_{audio}$ on a global level. %
 Specifically, we map $\mu(\mathbf{Z}_{face})$ and $\sigma(\mathbf{Z}_{face})$ to $\mu_{predict}$ and $\sigma_{predict}$ with a linear layer, where $\mu_{predict}$ and $\sigma_{predict}$ have the same channel number as $\mathbf{Z}_{audio}$. Finally, we adjust $\mathbf{Z}_{audio}$ with a similar manner as AdaIN~\cite{huang2017arbitrary}:
\begin{equation}
    \mathbf{Z}_{fuse} = \sigma_{predict}(\frac{\mathbf{Z}_{audio} - \mu(\mathbf{Z}_{audio})}{\sigma(\mathbf{Z}_{audio})}) + \mu_{predict}.
\end{equation}
The acquired $\mathbf{Z}_{fuse}$ contains both speech driven and speech independent information. %

\subsection{Sparsity Regularizer}
\label{sec:sparse}

When we linearly embed $\mathrm{Norm}(\mathbf{S}_{i})$ to low-dimensional facial feature vectors with embedding matrix $\mathbf{W}$, %
it is necessary to consider the regional nature of the facial movements, or the resulting feature vectors may fail to capture the subtle details in local regions. %

A simple and straightforward method for incorporating regionality is to divide the face into several regions according to face anatomy and apply a separate embedding matrix to each region. %
In this way, each region can be embedded independently and with greater detail. %
However, this approach can be labor-intensive as it requires the manual splitting of 3D faces into different regions, which is time-consuming and require expert knowledge. %
Alternatively, some previous methods~\cite{cudeiro2019capture} also initializes $\mathbf{W}$ from parametric 3D face model~\cite{FLAME:SiggraphAsia2017}. %
Such initialization does help the model to capture better facial details, but it can only synthesize 3D face mesh which has the same topology as the parametric 3D face model. %
When we switch the 3D face template, this method fails. %

Our proposed approach aims to address the issue of regional nature by leveraging a novel and efficient strategy, which does not require manual labor and is applicable to different 3D face templates. To achieve this, we utilize a sparse regularization technique inspired by Lasso Regression~\cite{tibshirani1996regression}. %
In particular, we apply \(\ell_1\) regularization to the embedding matrix $\mathbf{W}$, %
which encourages several elements of the weight matrix to be close to zero, resulting in sparsity. %
The sparsity enables each element of the feature vector to focus on the local facial regions. %
Furthermore, this strategy also improves the interpretability of the learned weights and leads to better generalization capability. %

\subsection{Backbone}


Designing an efficient and effective backbone is also crucial for the task of audio-driven 3D face animation. In this section, we  illustrate how the backbone obtains the latent audio feature $\mathbf{Z}_{audio}$ and how the backbone generates $\{\mathbf{\hat{S}}_{1} \cdots \mathbf{\hat{S}}_{T}\}$ from $\mathbf{Z}_{fuse}$ and $\mathbf{\bar{S}}$. %

We utilize the pre-trained HuBERT model~\cite{hsu2021hubert} for audio encoding. Notice that we have compared HuBERT, wav2vec 2.0~\cite{baevski2020wav2vec}, Mel spectrogram, and DeepSpeech~\cite{amodei2016deep} features. Among these features, we observe that the HuBERT feature performs best. The HuBERT model is a self-supervised method for learning audio representations that achieves state-of-the-art performance on various downstream tasks. The model is designed to take raw audio waveforms as input and generate a sequence of high-level representations that capture various aspects of the audio signal. In our implementation, we extract the final layer of the HuBERT model and resample the output with the desired frame rate to obtain the latent audio feature $\mathbf{Z}_{audio}$. %
During the training process, we do not fix the HuBERT model as the previous method~\cite{fan2022faceformer} does. %
Instead, we adopt a warm-up strategy. %
More specifically, at the start of training, we fix the HuBERT model and train the other sub-modules. Once the other sub-modules almost converge, we unfreeze the HuBERT model to allow it to fine-tune the task. %
Such a strategy has the advantage of preventing the scratch-initialized sub-modules from disturbing the pre-trained HuBERT model, which already contains useful and high-quality audio representations. %
With the warm-up strategy, we achieve faster convergence and better performance compared to simply freezing the pretrained audio model. %

For the other sub-modules in our backbone, we extensively employ the ResNet1D~\cite{he2016deep} structure rather than the Transformer~\cite{vaswani2017attention} structure. %
The ResNet1D conducts 1D convolution on the input feature vector sequence along the temporal dimension. %
It has the following three characteristics: %
(1) ResNet1D imposes a strong inductive bias on the model architecture, which aggregates information from temporal-adjacent frames. Such inductive bias lessens the required data for training. %
(2) ResNet1D has a strong capability for non-linear translation due to the stack of numerous convolution layers. %
(3) ResNet1D is a non-autoregressive and fully convolutional architecture, thereby it requires less computation cost and adapts to input with arbitrary size. %
These characteristics tally well with the task of speech-driven 3D face animation. %
Usually, the input speech and 3D facial animation  have strict temporal correspondence, %
such property has lessened the requirement of building complex time dependencies. %
When mapping speech to 3D facial animation, it is sufficient to fuse temporal information from adjacent frames as the ResNet1D does. %
Moreover, the input speech and 3D facial animation are highly heterogeneous, therefore the powerful non-linear translation capacity of ResNet1D is in need for our task. %

Based on the intuition above, we generate $\{\mathbf{\hat{S}}_{1} \cdots \mathbf{\hat{S}}_{T}\}$ from $\mathbf{Z}_{fuse}$ and $\mathbf{\bar{S}}$ with the ResNet1D decoder. %
We first embed $\mathbf{\bar{S}}$ with the embedding matrix $\mathbf{W}$ mentioned in Section~\ref{sec:sparse}, and then add the embedded vector to each frame of $\mathbf{Z}_{fuse}$. %
Afterward, the ResNet1D decoder takes the added embedding as input, and outputs the predicted movement features $\mathbf{Z}_{pred}$ with shape $T \times C_{face}$. %
Based on $\mathbf{Z}_{pred}$, we synthesize the 3D face sequences with the following equation:
\begin{equation}
    \mathbf{\hat{S}_{i}} = \mathbf{\bar{S}} + \alpha \mathbf{W}^{T}\times\mathbf{Z}_{pred}[i],
\end{equation}
where $\mathbf{W}^{T}$ is the transposed matrix of the embedding matrix $\mathbf{W}$. %
We scale the predicted movements with a coefficient $\alpha$ for faster convergence, 
 $\alpha$ is set to 0.1 in our implementation. %
Notice that we have removed all of the downsampling layers in ResNet1D during the decoding process; all layers have a convolutional kernel with size 3 and stride 1. %
Such modification avoids synthesizing over-smooth animations and retains high-frequency details. %

Overall, both the encoder and the decoder of our backbone are non-autoregressive, %
thus can be trained in parallel and run efficiently during inference. The non-autoregressive design also allows for flexible and variable-length input sequences, making our model applicable to various applications. 

\textbf{Training objectives.} We simultaneously optimize the \(\ell_2\) loss and the sparsity regularization loss. %
The \(\ell_2\) loss calculates the distance between the synthesized 3D face sequences $\{\mathbf{\hat{S}}_{1} \cdots \mathbf{\hat{S}}_{T}\}$ and the ground truth 3D face sequences $\{\mathbf{S}_{1} \cdots \mathbf{S}_{T}\}$. %
The sparsity regularization loss minimizes the \(\ell_1\) norm of $\mathbf{W}$. %
Formally:
\begin{equation}
\begin{split}
    \mathcal{L} & = \mathcal{L}_{\mathrm{\ell_{2}}} + \beta\mathcal{L}_{\mathrm{reg}} = \sum_{i = 1}^{T}||\mathbf{\hat{S}}_{i} - \mathbf{S}_{i}||_{2} + \beta||\mathbf{W}||_{1}.
\end{split}
\end{equation}

\subsection{Implementation Details}

For the HuBERT model, we adopt the HuBERT-large configuration with 24 Transformer layers. %
For the ResNet1D model, we adopt the ResNet18 configuration. %
The mesh embedding matrix $\mathbf{W}$ has a shape of $3N \times 256$, where $N$ is the number of mesh vertices. %
Different datasets have different numbers of mesh vertices. %
During training, we leverage the Adam optimizer~\cite{DBLP:journals/corr/KingmaB14} with learning rate of $10^{-4}$. %
The weight decay of the Adam optimizer is set to $0$ because it contradicts our sparsity regularizer. %
We train for 120 epochs with a mini-batch size of 8 samples. %
In the implementation, the scaling coefficient $\beta$ of the regularization loss is set to $10^{-4}$. %
We fix the HuBERT model at the first ten training epochs and unfreeze it at the subsequent epochs. %
We leverage one RTX3090 GPU for training. The training process takes less than one hour. %

\begin{figure*}[h]
  \centering
  \includegraphics[width=\linewidth]{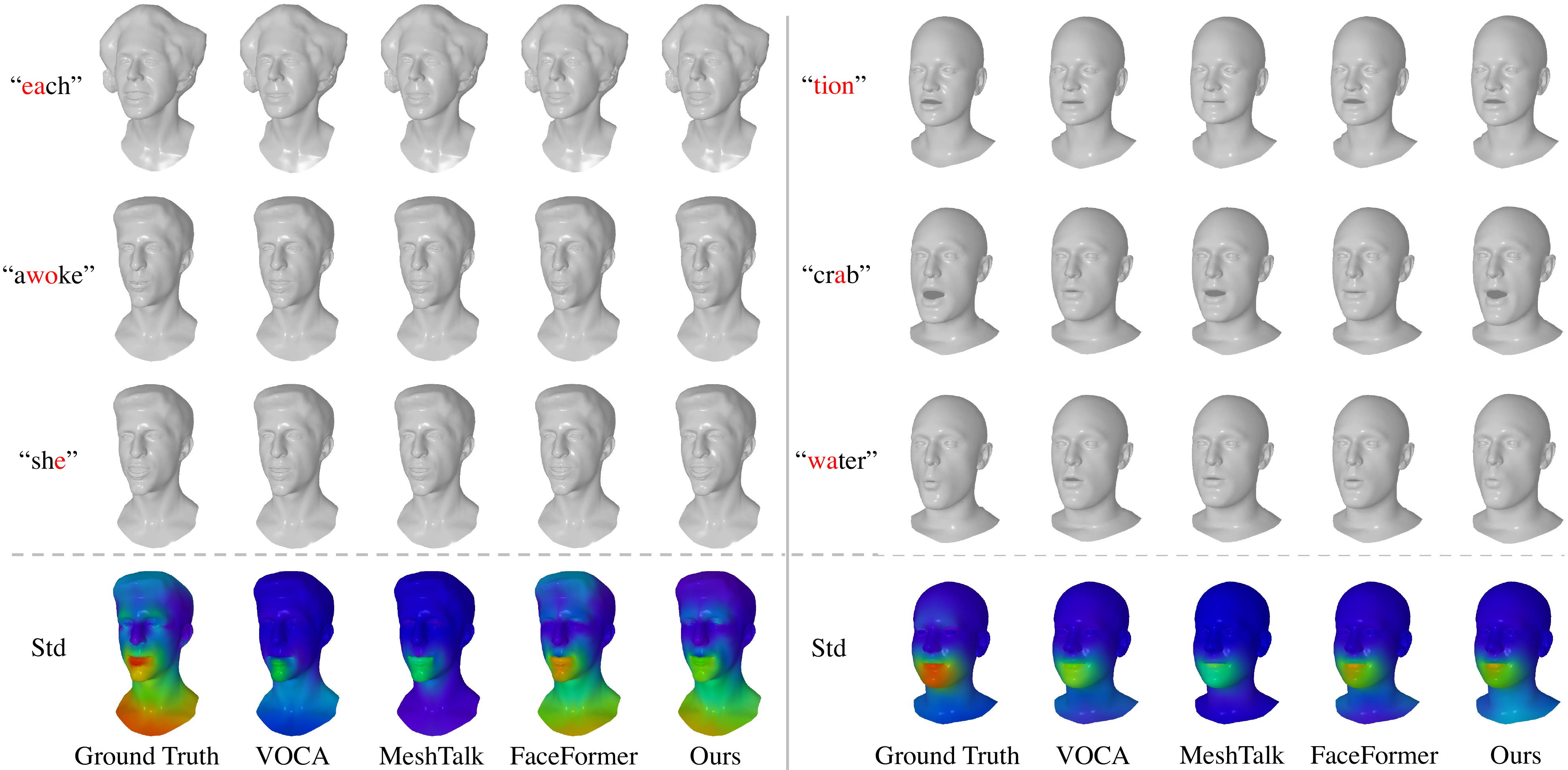}
  \caption{Qualitative comparison with baseline methods on MeshTalk dataset (left) and VOCASET (right). The first three rows show the facial animations when speaking different phonemes. The bottom row shows the standard deviation of facial animations; red denotes a large standard deviation, while blue denotes a smaller one. }
  \label{fig:qualitative}
\end{figure*}

\section{Experiments}

\subsection{Experimental Settings}

\indent\textbf{VOCASET dataset}~\cite{cudeiro2019capture}. %
VOCASET contains 480 3D face sequences obtained from 12 individuals. %
The 3D face mesh of each sequence adopts the template of the FLAME face model~\cite{FLAME:SiggraphAsia2017} with 5023 vertices.
We adopt the same training, validation, and testing splits for fair comparisons as VOCA~\cite{cudeiro2019capture}. Eight individuals are selected for training, two individuals are selected for validation, and two individuals are selected for testing. %

\textbf{BIWI dataset}~\cite{fanelli20103}. %
The BIWI dataset contains expressive emotions. %
Each mesh of the BIWI dataset contains 23370 vertices. %
We adopt the same evaluation protocol as FaceFormer~\cite{fan2022faceformer} on the BIWI dataset. %
More specifically, we exclude the neutral sentences from BIWI during evaluation and select only the emotional sentences. %
The training set has 192 sentences, the validation set has 24 sentences, and the testing set has 32 sentences. %

\textbf{MeshTalk dataset}~\cite{richard2021meshtalk}. %
The MeskTalk dataset is not fully open-sourced. %
Among all of the 250 individuals in the dataset, only the 3D face animations of 13 individuals are publicly available. %
Each mesh of the MeskTalk dataset contains 6172 vertices. %
Among the 13 individuals, we selected nine individuals for training, four for validation, and two for testing. The testing and validation sets overlap partially in individuals. %


\textbf{Baseline Methods}. %
We conducted a comparative evaluation of our proposed framework with three state-of-the-art methods: VOCA~\cite{cudeiro2019capture}, MeshTalk~\cite{richard2021meshtalk}, and FaceFormer~\cite{fan2022faceformer}. While VOCA and FaceFormer methods are conditioned on the identity label and driven speech, the MeshTalk method is conditioned on the 3D face sequence and driven speech. To ensure a fair comparison between the baseline and our proposed methods, we adopted the evaluation protocols of the respective methods. For the evaluation of VOCA and FaceFormer methods, we synthesized 3D face sequences based on test speech and training identities following the evaluation protocol of FaceFormer. For the evaluation of the MeshTalk method and our approach, we follow the evaluation protocol of MeshTalk, which obtains 3D face sequences from test speech and test 3D face sequences. During the evaluation process, we took great care to ensure no information leakage between the synthesized 3D face sequences and the testing 3D face sequences. The evaluation metrics were computed between the synthesized 3D face sequences and the testing 3D face sequences.

\subsection{Quantitative Evaluations}

We quantitatively evaluate the synchronization between the driven audio and the synthesized 3D face animations. %
To evaluate lip synchronization, we adopted two metrics: maximal lip vertex error~($L_{\max}^{lip}$) and average lip vertex error~($L_{\mathrm{mean}}^{lip}$). $L_{\max}^{lip}$ firstly computes the maximum Euclidean distance of lip region vertices between the synthesized and the ground truth 3D face and then averages the error among frames. %
$L_{\mathrm{mean}}^{lip}$ calculates the average distance between lip region vertices. %
To evaluate the synchronization of speech-independent movements, we utilize the following metrics: %
average upper face error~($L_{\mathrm{mean}}^{upper}$) and average face error~($L_{\mathrm{mean}}^{face}$). %
$L_{\mathrm{mean}}^{upper}$ and $L_{\mathrm{mean}}^{face}$ respectively calculate the average Euclidean distance of the upper and the whole face. %
Additionally, we calculate face dynamic time wrapping error~(F-DTW), which compares the temporal dynamics of the synthesized and ground truth 3D face sequences. F-DTW measures the similarity of two temporal sequences by finding an optimal warping path to align the sequences in time. %

\begin{table*}[]
\setlength\tabcolsep{0.8pt}
\centering
\caption{Comparison with state-of-the-art methods, lower denotes better for all metrics. Our method outperforms baseline methods in terms of both speech-driven movements and speech-independent movements. Notice that the scaling of metrics across different datasets is inconsistent due to variations in the scales of the original data.}
\begin{tabular}{@{}c||ccccc||ccccc||ccccc@{}}
\toprule
Dataset    & \multicolumn{5}{c||}{VOCASET~\cite{cudeiro2019capture}}                     & \multicolumn{5}{c||}{MeshTalk dataset~\cite{richard2021meshtalk}}            & \multicolumn{5}{c}{BIWI dataset~\cite{fanelli20103}}                         \\ \midrule
Method     & $L_{\mathrm{mean}}^{lip}$ & $L_{\max}^{lip}$ & $L_{\mathrm{mean}}^{upper}$ & $L_{\mathrm{mean}}^{face}$ & F-DTW & $L_{\mathrm{mean}}^{lip}$ & $L_{\max}^{lip}$ & $L_{\mathrm{mean}}^{upper}$ & $L_{\mathrm{mean}}^{face}$ & F-DTW & $L_{\mathrm{mean}}^{lip}$ & $L_{\max}^{lip}$ & $L_{\mathrm{mean}}^{upper}$ & $L_{\mathrm{mean}}^{face}$ & F-DTW \\ \midrule 
VOCA~\cite{cudeiro2019capture}       & 0.00324  & 0.00630 & 0.00054 & 0.00091   & 0.207 & 2.778    & 4.968   & 0.717   & 1.323     & 135.5 & 0.0235   & 0.0429  & 0.0089  & 0.0136    & 1.948 \\
MeshTalk~\cite{richard2021meshtalk}   & 0.00350  & 0.00640 & 0.00055 & 0.00092   & 0.210 & 2.516    & 4.556   & 0.776   & 1.268     & 129.3 & 0.0227   & 0.0424  & 0.0082  & 0.0126    & 1.779 \\
FaceFormer~\cite{fan2022faceformer} & 0.00212  & \textbf{0.00438} & 0.00046 & 0.00077   & \textbf{0.091} & 2.206    & 3.885   & 0.711   & 1.210     & 123.9 & 0.0230   & 0.0402  & 0.0092  & 0.0143    & 2.047 \\ \midrule
Ours w/o Composite       & 0.00171  & 0.00470 & \textbf{0.00041} & 0.00064   & 0.145 & 1.728    & 3.474   & 0.631   & 0.938     & 96.6  & 0.0186   & 0.0381  & 0.0071  & 0.0120    & 1.491 \\
Ours w/o Regional       & 0.00166  & 0.00451 & \textbf{0.00041} & \textbf{0.00063}   & 0.142 & 1.690    & 3.416   & 0.632   & \textbf{0.927}     & \textbf{95.6}  & 0.0175   & 0.0366  & 0.0069  & 0.0107    & 1.376 \\
\textbf{Ours}       & \textbf{0.00161}  & 0.00447 & 0.00042 & 0.00065   & 0.147 & \textbf{1.659}    & \textbf{3.382}   & \textbf{0.621}   & 0.930     & 96.2  & \textbf{0.0170}   & \textbf{0.0353}  & \textbf{0.0068}  & \textbf{0.0105}    & \textbf{1.330} \\
\bottomrule
\end{tabular}
\label{tab:baseline}
\end{table*}


Table~\ref{tab:baseline} presents the comparison results among the methods. Notably, the MeshTalk method shows suboptimal performance due to its original implementation's heavy reliance on large-scale training data, which is not available in our experimental setup. As a result, the MeshTalk method exhibits inadequate generalization capacity when the training data only comprises around ten individuals. %
In contrast, the FaceFormer and our methods have better generalization capacity due to the incorporation of pretrained audio models~\cite{baevski2020wav2vec,hsu2021hubert}. Furthermore, our method outperforms the FaceFormer method by a large margin in terms of lip synchronization and speech-independent synchronization.

In addition, our method is also computationally efficient due to the design of non-autoregressive architecture. %
Our method takes 0.007 seconds to synthesize 1-second 3D face sequences during inference, while the FaceFormer method takes 0.1 seconds. %
The efficiency of our method makes it practical for real-time applications such as video conferencing, telepresence, and gaming. %


\subsection{Qualitative Evaluations}


\begin{figure}[h]
    \centering
    \includegraphics[width=\linewidth]{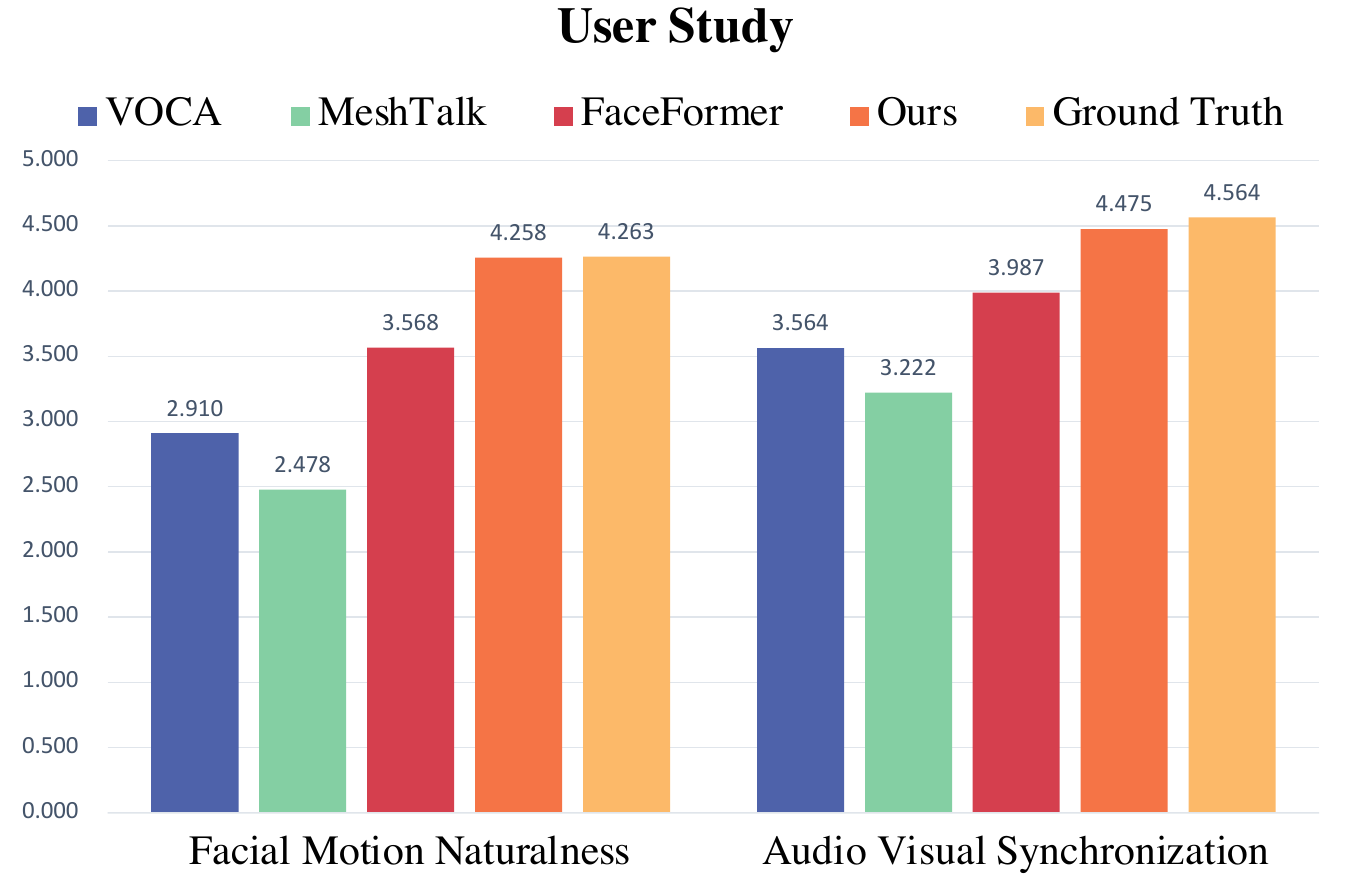}
    \caption{User study results. The average scores of facial motion naturalness and audio-visual synchronization are reported. Higher scores denote better results.}
    \label{fig:user}
\end{figure}

We qualitatively compare our method with baseline methods in Figure~\ref{fig:qualitative}. %
The first three rows give the synthesized 3D faces for different input speeches. %
For fair comparisons, all methods and input speeches are required to utilize the same talking style. %
Our method generates more realistic and vivid 3D facial movements with better lip synchronization than the baseline methods. %
Mainly, our method exhibits more significant mouth opening and closing movements for pronouncing /b/ and /p/ and more evident pouting movements for pronouncing /w/ and /v/.
Meanwhile, the baseline methods affect jaw flapping, resulting in unnatural facial movements. %

The bottom row demonstrates that our and FaceFormer methods synthesize diversified speech-independent facial movements. %
We generate a heat map to visualize the intensity of facial movements, with vertices that exhibit a larger distance of motion appearing in red. In comparison, those with a shorter distance appear in blue. %
We observe that our and FaceFormer methods synthesize more intense movements for the lip region and the other facial regions. In contrast, the VOCA and MeshTalk methods tend to synthesize over smooth movements with little variation in intensity. %
By generating more diverse and intense facial movements, our method can create more vivid and expressive 3D faces, enhancing the overall realism of the synthesized output. %

Additionally, we conduct user studies to evaluate the naturalness of facial motion and audio-visual synchronization. Specifically, we randomly select eight audio samples to drive the 3D face animations. We synthesize the 3D face animations with the same talking style as the input audio for each method. We invite 14 participants to rate the facial motion naturalness and audio-visual synchronization. The participants are asked to provide ratings on two aspects: (1) whether the facial motion appears natural and (2) whether the audio and 3D face animation are synchronized properly. %
Participants rate the mean opinion score (MOS)~\cite{recommendation2006vocabulary} using a 1-5 scale, with higher scores indicating better results. %
Figure~\ref{fig:user} reports the results of user studies. %
We received feedback from several participants indicating that VOCA and MeshTalk methods exhibit unnatural facial movements due to the jaw-flapping effect. %
Compared with the FaceFormer method, our method synthesizes facial movements with more subtle details. %
Overall, these results highlight the effectiveness of our proposed method in generating high-quality 3D facial animations. %

\subsection{Ablation Study}

\begin{figure}[h]
  \centering
  \includegraphics[width=\linewidth]{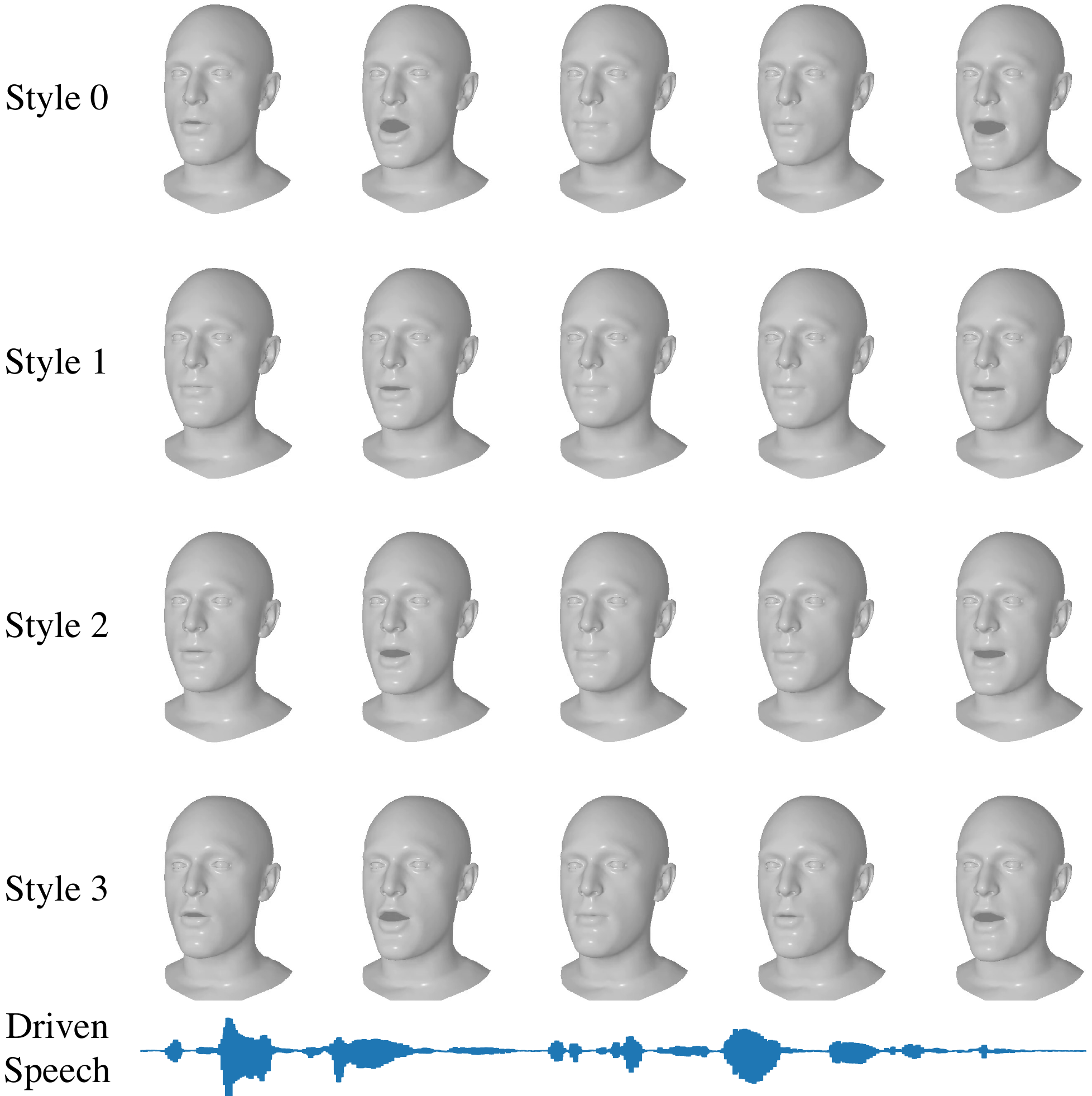}
  \caption{Visualizations of stylized 3D facial animations driven by input speech. }
  \label{fig:style}
\end{figure}

We conducted ablation studies to evaluate the effectiveness of considering composite and regional natures in synthesizing speech-driven 3D face animations. Specifically, we performed experiments by selectively removing the composite or regional nature from our model architecture and synthesizing speech-driven 3D face animations using the modified models. The quantitative results of these experiments are reported in Table~\ref{tab:baseline}. %

The results of the MeshTalk and BIWI datasets reveal that removing the regional nature from our model has a significant negative impact on the performance of lip synchronization. This is because the regional nature allowed our model to focus on specific details of facial movements around the lips, which are crucial for accurately synchronizing lip movements with speech. On the other hand, when we remove the composite nature from our model, we observed a decline in performance for synthesizing speech-independent facial movements. This is because the composite nature enabled our model to capture global patterns and general trends in facial movements that are independent of speech. It is worth noting that the VOCASET dataset has minimal speech-independent facial movements, and hence, the performance inside the ablation study was similar. Our findings suggest that composite and regional natures are important for synthesizing speech-driven 3D face animations.

\begin{figure}[h]
  \centering
  \includegraphics[width=\linewidth]{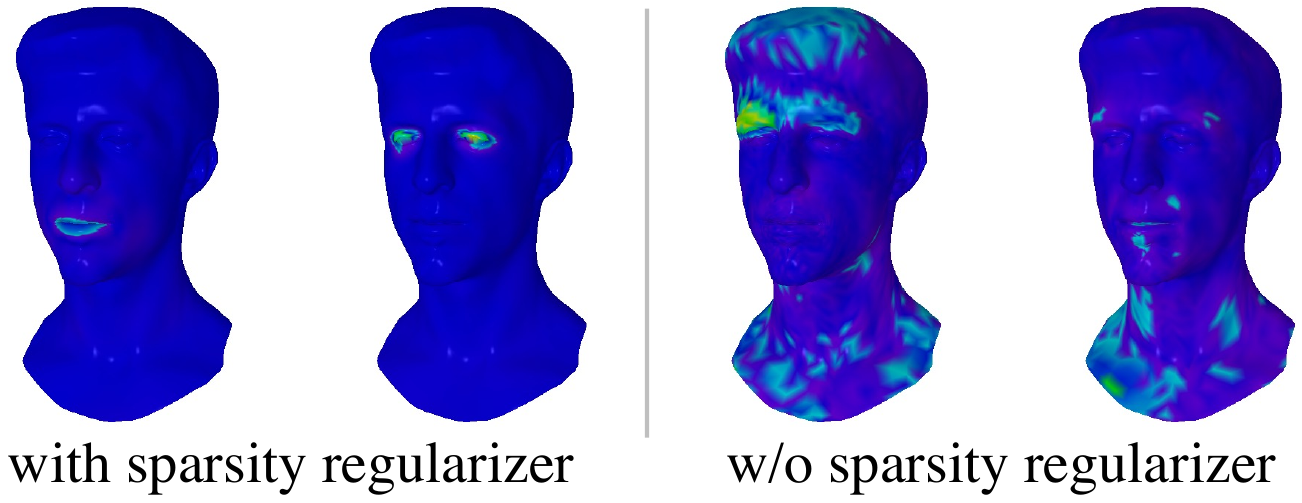}
  \caption{The activated region for each element of the motion features. Blue regions indicate no influence, while red regions denote high influence, and green regions represent a transitional influence between blue and red. }
  \label{fig:region}
\end{figure}

We have also created visualizations to showcase the effectiveness of our method in synthesizing composite and regional facial movements. Figure~\ref{fig:style} shows that our method successfully generates diverse talking styles for single-driven speech input, highlighting the ability to effectively capture the nuances and variations in facial expressions characteristic of different speaking styles. %
To further demonstrate the effectiveness of our method, we also draw Figure~\ref{fig:region} to illustrate the impact of our proposed sparsity regularizer, which enforces each facial feature element to focus on the local region of mesh vertices. By incorporating this sparsity regularizer, our method can identify and extract interpretable regions for synthesizing facial movements, leading to more natural and accurate results. Meanwhile, when the sparsity regularizer is removed, the activated regions of motion features spread across the face. %

Overall, our ablation study provides empirical evidence for the effectiveness of considering both composite and regional natures in synthesizing speech-driven 3D face animations. By combining these two natures, our model can achieve superior performance.


\section{Conclusion}

This paper emphasizes the importance of considering composite and regional natures in speech-driven 3D face animation. We conducted extensive observations demonstrating that these natures are prevalent in 3D facial movements. Our proposed comprehensive 3D face animation framework incorporates both of these natures. To handle the composite nature, we introduced an adaptive modulating module that extracts speech-independent information from arbitrary 3D face sequences and fuses this information with the driving audio. To address the regional nature, we proposed a sparsity regularizer that enforces each element of the motion feature to focus on local regions of 3D faces. Furthermore, we designed an efficient non-autoregressive backbone for mapping audio and 3D facial movements. Our backbone is built on a pretrained HuBERT model and a ResNet1D network, which preserves high-frequency details of facial movements. During implementation, our backbone synthesizes one second of facial animations with 30 fps in only 0.007 seconds. Extensive experiments demonstrate that our proposed framework outperforms baseline methods both quantitatively and qualitatively, with significantly reduced computational cost.

\section{Acknowledgements}
This work is supported by the National Key R\&D Program of China under Grant No. 2020AAA0108600, the State Key Program of the National Natural Science Foundation of China (NSFC) (No.61831022), and supported in part by the Natural Science Foundation of China under Grant No.62222606 and 62076238.

\newpage
\bibliographystyle{ACM-Reference-Format}
\balance
\bibliography{sample-base}

\end{document}